\documentclass[10pt]{article}

\usepackage{amsmath,amsthm,verbatim,amssymb,amsfonts,amscd, graphicx}
\usepackage{graphics}
\usepackage{centernot}
\usepackage{authblk}
\theoremstyle{plain}
\newtheorem{theorem}{Theorem}

\newtheorem{lemma}{Lemma}
\newtheorem*{remark}{Remark}

\newtheorem{definition}{Definition}

\usepackage[colorlinks,linkcolor=blue,citecolor=blue]{hyperref}

\usepackage[bottom]{footmisc}
\usepackage{caption}
\usepackage{subcaption}
\usepackage{helvet}  
\usepackage{courier}  
\usepackage{url}  
\usepackage{graphicx}  
\usepackage{multirow}
\usepackage{amsthm}
\usepackage{color}
\usepackage{MnSymbol}
\usepackage{makecell}
\usepackage{arydshln}
\usepackage{amsmath}
\usepackage{caption} 
\usepackage{natbib}

\usepackage{textcomp}
\usepackage{wrapfig}
\usepackage{algorithm}
\usepackage{algorithmic}

\usepackage{csquotes}

\newcommand{\E}{\mathbb{E}}

\begin{document}

\title{Exact Phase Transitions in Deep Learning}
\author{Liu Ziyin$^1$, Masahito Ueda$^{1,2,3}$\\
{\small ${}^1$\textit{Department of Physics, The University of Tokyo, 7-3-1 Hongo, Bunkyo-ku, Tokyo 113-0033}\\
${}^2$\textit{Institute for Physics of Intelligence, The University of Tokyo, 7-3-1 Hongo, Bunkyo-ku, Tokyo 113-0033}\\
${}^3$\textit{RIKEN Center for Emergent Matter Science (CEMS), Wako, Saitama 351-0198, Japan}}}
\maketitle

\begin{abstract}
    This work reports deep-learning-unique first-order and second-order phase transitions, whose phenomenology closely follows that in statistical physics. In particular, we prove that the competition between prediction error and model complexity in the training loss leads to the second-order phase transition for nets with one hidden layer and the first-order phase transition for nets with more than one hidden layer. The proposed theory is directly relevant to the optimization of neural networks and points to an origin of the posterior collapse problem in Bayesian deep learning.
\end{abstract}

Understanding neural networks is a fundamental problem in both theoretical deep learning and neuroscience. In deep learning, learning proceeds as the parameters of different layers become correlated so that the model responds to an input in a meaningful way. This is reminiscent of an ordered phase in physics, where the microscopic degrees of freedom behave collectively and coherently. Meanwhile, regularization effectively prevents the overfitting of the model by reducing the correlation between model parameters in a manner similar to the effect of an entropic force in physics. One thus expects a phase transition in the model behavior from the regime where the regularization is negligible to the regime where it is dominant. In the long history of statistical physics of learning \cite{hopfield1982neural, watkin1993statistical, martin2017rethinking, bahri2020statistical}, a series of works studied the under-to-overparametrization (UO) phase transition in the context of linear regression \cite{krogh1992generalization, krogh1992simple, watkin1993statistical, haussler1996rigorous}.
Recently, this type of phase transition has seen a resurgence of interest \cite{hastie2019surprises, liao2020random}. One recent work by \cite{li2021statistical} deals with the UO transition in a deep linear model. However, the UO phase transition is not unique to deep learning because it appears in both shallow and deep models and also in non-neural-network models \cite{belkin2020two}. To understand deep learning, we need to identify what is unique about deep neural networks.

In this work, we address the fundamental problem of the loss landscape of a deep neural network and prove that there exist phase transitions in deep learning that can be described precisely as the first- and second-order phase transitions with a striking similarity to physics. We argue that these phase transitions can have profound implications for deep learning, such as the importance of symmetry breaking for learning and the qualitative difference between shallow and deep architectures. We also show that these phase transitions are unique to machine learning and deep learning. They are unique to machine learning because they are caused by the competition between the need to make predictions more accurate and the need to make the model simpler. These phase transitions are also deep-learning unique because they only appear in ``deeper" models. For a multilayer linear net with stochastic neurons and trained with $L_2$ regularization,
\begin{enumerate}
    \item we identify an order parameter and effective landscape that describe the phase transition between a trivial phase and a feature learning phase as the $L_2$ regularization hyperparameter is changed (Theorem~\ref{theo: existence of order parameter});
    \item we show that finite-depth networks cannot have the zeroth-order phase transition (Theorem~\ref{theo: no zeroth order});
    \item we prove that:
    \begin{enumerate}
        \item depth-$0$ nets (linear regression) do not have a phase transition (Theorem~\ref{theo: depth 0});
        \item depth-$1$ nets have the second-order phase transitions (Theorem~\ref{theo: depth 1});
        \item depth-$D$ nets have the first-order phase transition (Theorem~\ref{theo: depth 1+}) for $D>1$;
        \item infinite-depth nets have the zeroth-order phase transition (Theorem~\ref{theo: infinite D}).
    \end{enumerate}
\end{enumerate}
The theorem statements and proofs are presented in the Supplementary Section~\ref{app: theory}. To the best of our knowledge, we are the first to identify second-order and first-order phase transitions in the context of deep learning. Our result implies that one can precisely classify the landscape of deep neural models according to the Ehrenfest classification of phase transitions.

\section*{Results}

\textbf{Formal framework}.  Let $\ell(w, a)$ be a differentiable loss function that is dependent on the model parameter $w$ and a hyperparameters $a$. The loss function $\ell$ can be decomposed into a data-dependent feature learning term $\ell_0$ and a data-independent term $a R(w)$ that regularizes the model at strength $a$:
\begin{equation}\label{eq: training loss decomposed}
    \ell(w,a) = \E_x[\ell_0(w,x)] + a R(w).
\end{equation}
Learning amounts to finding the global minimizer of the loss:
\begin{equation}
\begin{cases}
    L(a) := \min_w \ell(w,a);\\
    w_* := \arg\min_w \ell(w,a).
\end{cases}
\end{equation}
Naively, one expects $L(a)$ to change smoothly as we change $a$. If $L$ changes drastically or even discontinuously when one perturb $a$, it becomes hard to tune $a$ to optimize the model performance. Thus, that $L(a)$ is well-behaved is equivalent to that $a$ is an easy-to-tune hyperparameter. We are thus interested in the case where the tuning of $a$ is difficult, which occurs when a phase transition comes into play. 

It is standard to treat the first term in Eq.~\eqref{eq: training loss decomposed} as an energy. To formally identify the regularization term as an entropy, its coefficient must be proportional to the temperature:
\begin{equation}
    a R (w) = \frac{T}{2\sigma^2} R(w),
\end{equation}
where $\sigma^2$ controls the fluctuation of $w$ at zero temperature. We note that this identification is consistent with many previous works, where the term that encourages a lower model complexity is identified as an ``entropy" \cite{haussler1996rigorous, vapnik2006estimation, benedek1991learnability, friston2009free, li2021statistical}. In this view, learning is a balancing process between the learning error and the model complexity. Intuitively, one expects phase transitions to happen when one term starts to dominate the other, just like thermodynamic phase transitions that take place when the entropy term starts to dominate the energy.

In this setting, the partition function is $Z(a)= \int dw \exp[-\ell(w,a)/T]$. We consider a special limit of the partition function, where both $T$ and $2\sigma^2$ are made to vanish with their ratio held fixed at $T/2\sigma^2 = \gamma$. In this limit, one can find the free energy with the saddle point approximation, which is exact in the zero-temperature limit:
\begin{align}
    F(a) = \lim_{T\to 0,\ \sigma^2\to 0,\ T/2\sigma^2 = \gamma}-T\log \int dw \exp[-\ell(w,a)/T] = \min_w \ell(w, a).
\end{align}
We thus treat $L$ as the free energy.

\begin{definition}\label{def: phase transition}
    $L(a)$ is said to have the $n$th-order phase transition in $a$ at $a=a^*$ if $n$ is the smallest integer such that $\frac{d^n}{da^n} L(a)|_{a=a^*}$ is discontinuous.
\end{definition}

We formally define the order parameter and effective loss as follows.
\begin{definition}\label{def: order parameter}
    $b=b(w)\in \mathbb{R}$ is said to be an order parameter of $\ell(w,a)$ if there exists a function $\bar{\ell}$ such that for all $a$, $\min_w \bar{\ell}(b(w),a) = L(a)$, where $\bar{\ell}$ is said to be an effective loss function of $\ell$.
\end{definition}
In other words, an order parameter is a one-dimensional quantity whose minimization on $\bar{\ell}$ gives $L(a)$. The existence of an order parameter suggests that the original problem $\ell(w,a)$ can effectively be reduced to a low-dimensional problem that is much easier to understand. Physical examples are the average magnetization in the Ising model and the average density of molecules in a water-to-vapor phase transition. A dictionary of the corresponding concepts between physics and deep learning is given in Table~\ref{tab:dictionary}. 

Our theory deals with deep linear nets, the primary minimal model for deep learning. It is well-established that the landscape of a deep linear net can be used to understand that of nonlinear networks \cite{kawaguchi2016deep,hardt2016identity,laurent2018deep}. The most general type of deep linear nets, with $L_2$ regularization and stochastic neurons, has the following loss:
\begin{equation}\label{eq: intro loss}
    \underbrace{\E_x \E_{\epsilon^{(1)}, \epsilon^{(2)},...,\epsilon^{(D)}} \left(\sum_{i,i_1,i_2,...,i_D}^{d_0,d_0,d_0,...d_0} U_{i_D} \epsilon^{(D)}_{i_{D}}...\epsilon_{i_2}^{(2)} W_{i_2i_1}^{(2)}\epsilon_{i_1}^{(1)} W_{i_1i}^{(1)} x_i  - y\right)^2}_{L_0} + \underbrace{\gamma||U||_2^2 + \sum_{i=1}^D\gamma ||W^{(i)}||_F^2}_{ L_2\ {\rm reg.}},
\end{equation}
where $x$ is the input data, $y$ the label, $U$ and $W^{(i)}$ the model parameters, $D$ the network depth, $\epsilon$ the noise in the hidden layer (e.g., dropout), $d_0$ the width of the model, and $\gamma$ the weight decay strength. We build on the recent results established in \cite{ziyin2022exact}. Let $b:=||U||/d_0$. Ref.~\cite{ziyin2022exact} shows that all the global minima of Eq.~\eqref{eq: intro loss} must take the form $U=f(b)$ and $W_i =f_i(b)$, where $f$ and $f_i$ are explicit functions of the hyperparameters. Ref.~\cite{ziyin2022exact} further shows that there are two regimes of learning, where, for some range of $\gamma$, the global minimum is uniquely given by $b=0$, and for some other range of $\gamma$, some $b>0$ gives the global minimum. When $b=0$, the model outputs a constant $0$, and so this regime is called the ``trivial regime," and the regime where $b=0$ is not the global minimum is called the ``feature learning regime." In this work, we prove that the transition between these two regimes corresponds to a phase transition in the Ehrenfest sense (Definition 1), and therefore one can indeed refer to these two regimes as two different phases.

\begin{table}[]
    \centering
    \begin{minipage}{0.4\linewidth}
        \begin{tabular}{c|c}
    machine learning  & statistical physics \\
    \hline
    training loss   & free energy\\
    prediction error  & internal energy \\
    regularization  & negative entropy \\
    \end{tabular}
    \end{minipage}
    \begin{minipage}{0.4\linewidth}
        \begin{tabular}{c|c}
    learning process & symmetry breaking \\
    \hline
    norm of model ($b$) & order parameter \\
    feature learning regime & ordered phase \\
    trivial regime & disordered phase \\
    noise required for learning & latent heat  \\
    \end{tabular}
    \end{minipage}
    \caption{Left table: the correspondence between machine learning and statistical physics. Right table: the correspondence between a learning process and symmetry breaking.}
    \label{tab:dictionary}
\end{table}

\textbf{No-phase-transition theorems}. The first result we prove is that there is no phase transition in any hyperparameter ($\gamma,\ E[xx^T],\ E[xy],\ E[y^2]$) for a simple linear regression problem. In our terminology, this corresponds to the case of $D=0$. The fact that there is no phase transition in any of these hyperparameters means that the model's behavior is predictable as one tunes the hyperparameters. In the parlance of physics, a linear regressor operates within the linear-response regime. 

Theorem~\ref{theo: no zeroth order} shows that a finite-depth net cannot have zeroth-order phase transitions. This theorem can be seen as a worst-case guarantee: the training loss needs to change continuously as one changes the hyperparameter. We also stress that this general theorem applies to standard nonlinear networks as well. Indeed, if we only consider the global minimum of the training loss, the training loss cannot jump. However, in practice, one can often observe jumps because the gradient-based algorithms can be trapped in local minima. The following theory offers a direct explanation for this phenomenon.

\begin{figure}[t!]
    \centering
    \includegraphics[width=0.3\linewidth]{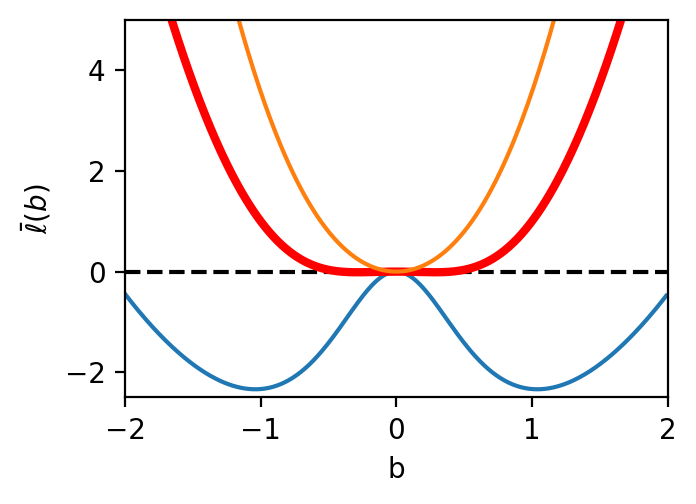}
    \includegraphics[width=0.32\linewidth]{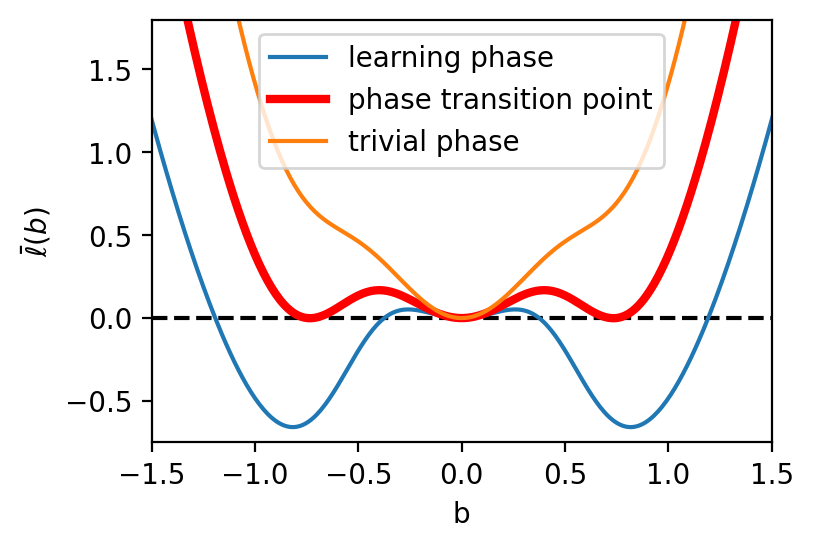}
    \vspace{-1em}
    \caption{Effective landscape given in Eq.~\eqref{eq: general effective loss} for $D=1$ (left) and $D=2$ (right). For $D=1$, zero is either the global minimum or a local maximum. Note that the shape of the loss resembles that of the Landau free energy for the second-order phase transition. For $D=2$, the landscape becomes more complicated, featuring the emergence of local minima. In particular, zero is always a local minimum.}
    \label{fig:effective loss}
\end{figure}

\textbf{Phase Transitions in Deeper Networks}. Theorem~\ref{theo: depth 1} shows that the quantity $b$ is an order parameter describing any phase transition induced by the weight decay parameter in Eq.~\eqref{eq: intro loss}. Let $b= ||U||/d_u$, $A_0:=\E[xx^T]$, and $a_i$ be the $i$-th eigenvalue of $A_0$. The effective loss landscape is
\begin{equation}\label{eq: general effective loss}
        \bar{\ell}(b, \gamma) := -\sum_i \frac{d_0^{2D}b^{2D} \E[x'y]_i^2 }{d_0^D (\sigma^2+d_0)^D a_i b^{2D} + \gamma}  + \E_x[y^2] +  \gamma D d_0^2 b^2,
\end{equation} 
where $x'$ is a rotation of $x$. See Figure~\ref{fig:effective loss} for an illustration. The complicated landscape for $D>1$ implies that neural networks are susceptible to initialization schemes and entrapment in meta-stable states is common (see Supplementary Section~\ref{app: sensitivity to the initial condition}).

Theorem~\ref{theo: depth 1+} shows that when $D=1$ in Eq.~\eqref{eq: intro loss}, there is a second-order phase transition precisely at \begin{equation}
    \gamma=||\E[xy]||.
\end{equation}
In machine learning language, $\gamma$ is the regularization strength  and $||E[xy]||$ is the signal. The phase transition occurs precisely when the regularization dominates the signal. In physics, $\gamma$ and $||\E[xy]||$ are proportional to the temperature and energy, respectively. The phase transition occurs exactly when the entropy dominates the energy. Also, the phase transition for a depth-$1$ linear net is independent of the number of parameters of the model. For $D>1$, the size of the model does play a role in influencing the phase transition. However, $\gamma$ remains the dominant variable controlling this phase transition. This independence of the model size is an advantage of the proposed theory because our result becomes directly relevant for all model sizes, not just the infinitely large ones that the previous works often adopt.

\begin{figure}[t!]
    \centering
    \includegraphics[width=0.325\linewidth]{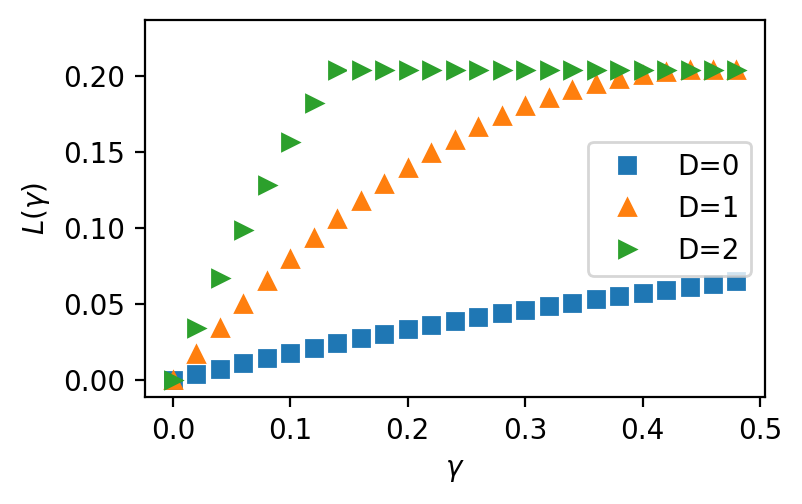}
    \includegraphics[width=0.325\linewidth]{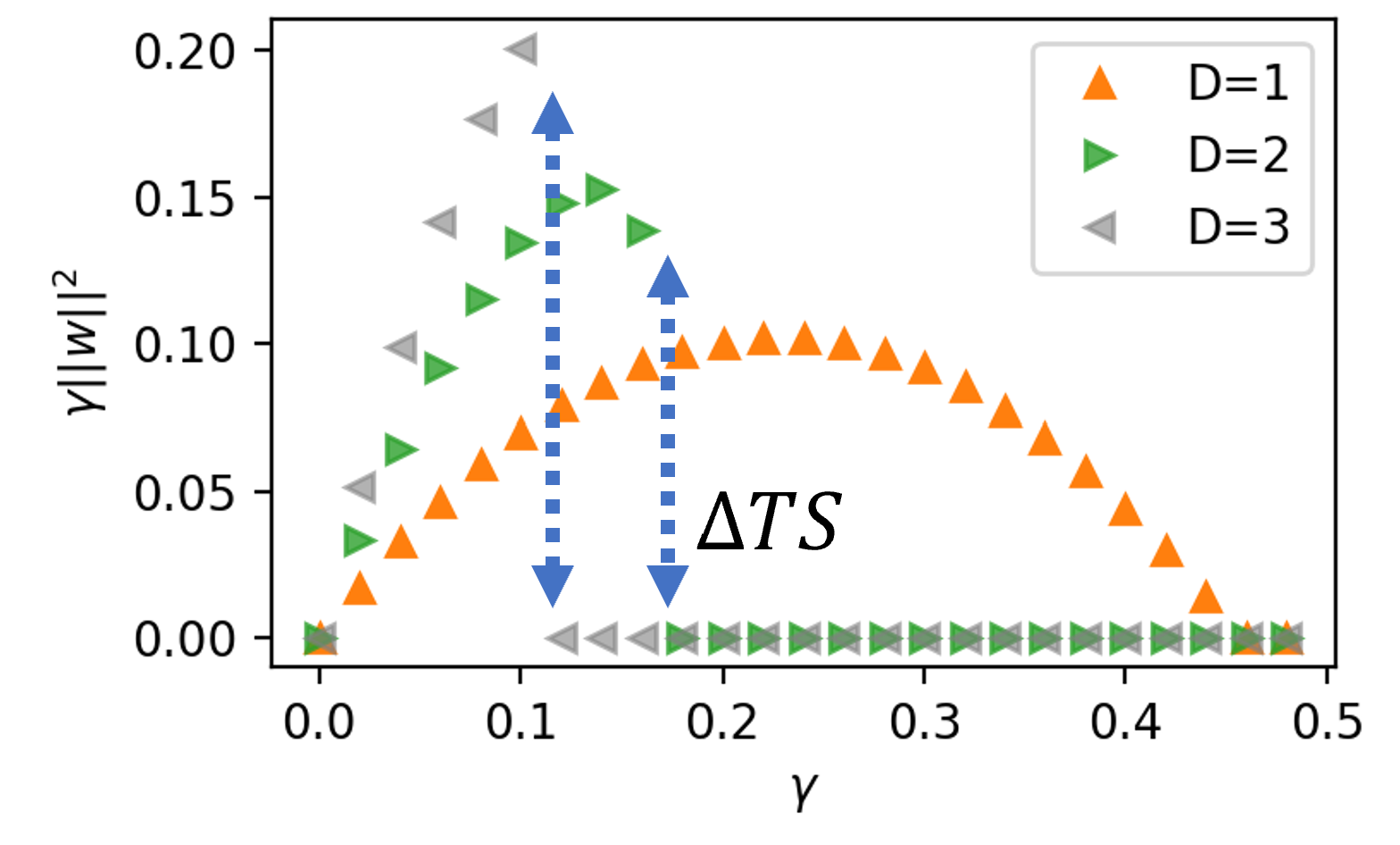}
    \includegraphics[width=0.325\linewidth]{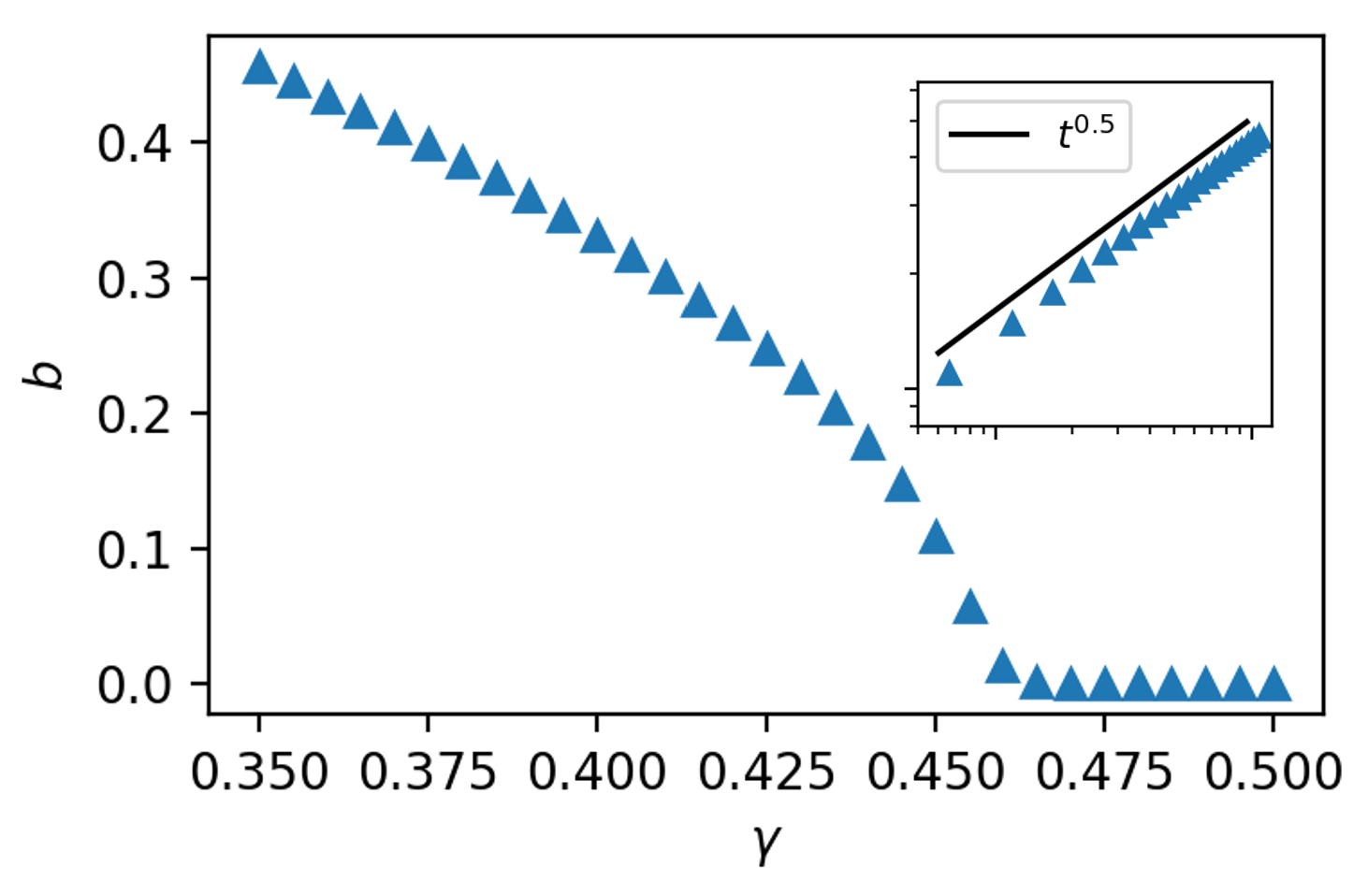}

    \vspace{-1em}
    \caption{Phase transitions in a linear net. In agreement with the theory, a depth-$0$ net has no phase transition. A depth-$1$ net has a second-order phase transition at approximately $\gamma = 0.45$, close to the theoretical value of $||\E[xy]||$, and a depth-$2$ net has a first-order phase transition at roughly $\gamma = 0.15$. The qualitative differences between networks of different depths are clearly observed in the data. \textbf{Left}: Training loss of a network with $0$ (linear regression), $1$, and $2$ hidden layers. Clearly, a depth-$0$ net shows no phase transition. A depth-$1$ net has a second-order phase transition at approximately $\gamma = 0.45$, and a depth-$2$ net has a first-order phase transition at roughly $\gamma = 0.15$. \textbf{Middle}: Magnitude of the regularization term at convergence. As discussed in the main text, this term corresponds to the entropy term $TS$. We see that for $D>1$, there is a jump (discontinuity) in $TS$ from a finite value to $0$. This jump corresponds to the latent heat of the first-order phase transition process. \textbf{Right}: Order parameter as a function of $\gamma$. The inset shows that $b$ precisely scales as $t^{0.5}$ with $t:=-(\gamma-\gamma^{*})$ in the vicinity of the phase transition, in agreement with the standard Landau theory. }
    \label{fig:phase transition}
    \vspace{1em}
 \end{figure}

For $D\geq 2$, we show that there is a first-order phase transition between the two phases at some $\gamma >0$. However, an analytical expression for the critical point is not known. In physics, first-order phase transitions are accompanied by latent heat. Our theory implies that this heat is equivalent to the amount of random noise we have to inject into the model parameters to escape from a local to the global minimum for a deep model. We illustrate the phase transitions studied in Figure~\ref{fig:phase transition}. We also experimentally demonstrate that the same phase transitions take place in deep nonlinear networks with the corresponding depths (Supplementary Section~\ref{app: nonlinear nets}). While infinite-depth networks are not used in practice, they are important from a theoretical point of view \cite{sonoda2019transport} because they can be used for understanding a (very) deep network that often appears in the deep learning practice. Our result shows that the limiting landscape has a zeroth-order phase transition at $\gamma=0$. In fact, zeroth-order phase transitions do not occur in physics, and it is a unique feature of deep learning.

\begin{figure}[t!]
    \centering
    \includegraphics[width=0.32\linewidth]{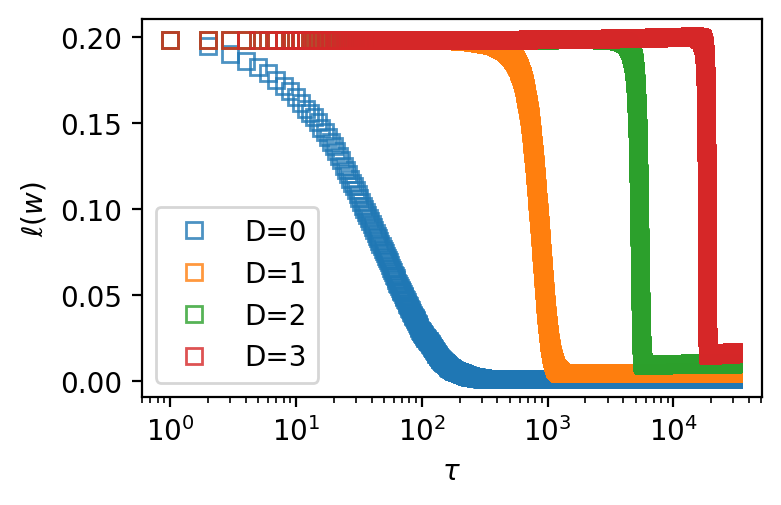}
    \includegraphics[width=0.325\linewidth]{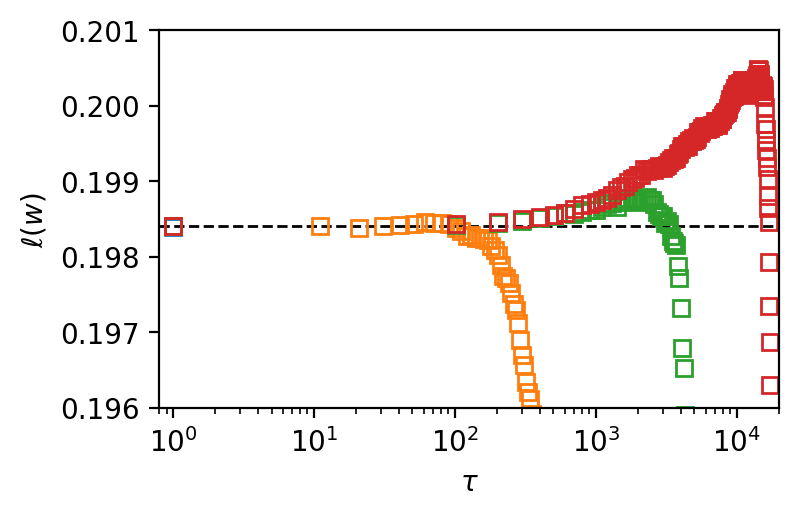}
    \includegraphics[width=0.32\linewidth]{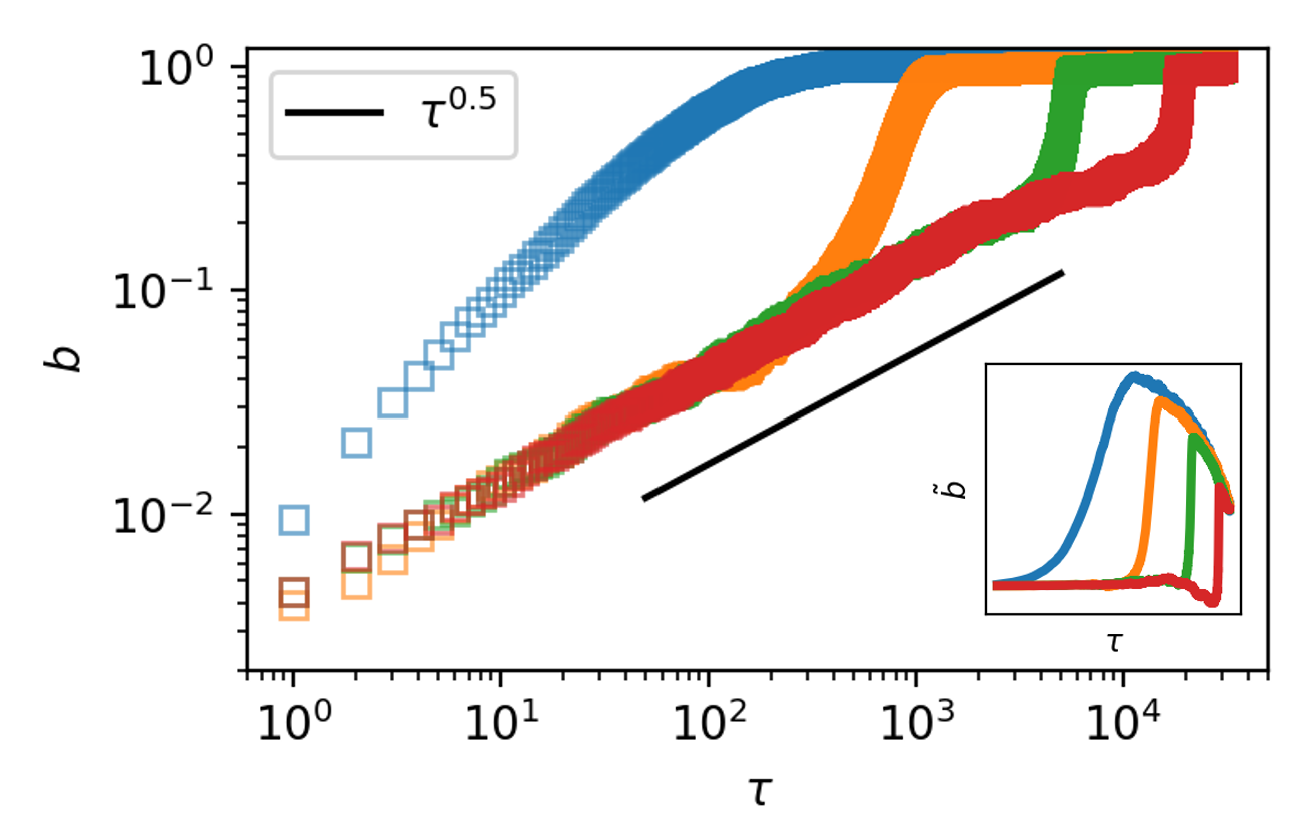}
    \caption{Dynamics of training, where the model is initialized at the origin and the learning proceeds under gradient descent with injected Gaussian noise. Before training, the models lie roughly in the trivial phase because the model is initialized close to the origin and has not learned anything yet. However, for the feature learning phase, any global minimum must choose a specific $b\neq 0$, and so the actual solution does not feature the symmetry in $b$: a symmetry breaking in $b$ must take place for the learning to happen. The recent work of Ref.~\cite{tanaka2021noether} showed that the symmetries in the loss could become difficult obstacles in the training of a neural network, and our result complements this view by identifying a precise deep-learning-relevant symmetry to be broken.  \textbf{Left}: the training loss $L$; except for $D=0$, where no symmetry breaking occurs, the dynamics exhibits a wide plateau that hinders learning emerging at initialization. \textbf{Middle}: a zoom-in of the left panel when $L$ is close to the initialized value ($\approx 0.2$).  For $D=1$, the loss decreases monotonically. For $D>1$, in sharp contrast, the loss first increases slowly and then decreases precipitously, a signature of escaping from a local minimum: the height of the peak may be interpreted as the latent heat of the phase transition since this is the "energy barrier" for the system to overcome in order to undergo the first-order phase transition. \textbf{Right}: time evolution of the order parameter $b$: one sees that $D=0$ shows a fast increase of $b$ from the beginning. For $D\geq 1$, the initial stage is dominated by slow diffusion, where $b$ increases as the square root of time. The diffusion phase only ends after a long period, before a fast learning period begins. One also notices that in the fast learning period, the slope of $b$ versus time is different for different depths, with deeper models considerably faster than shallower ones. The inset shows the corrected order parameter $\tilde{b}:= b - D\sqrt{\tau}$, where $\tau$ is the training step, and $D$ is the diffusion constant of the noisy gradient descent. One sees that $\tilde{b}$ stays zero over an extended period of time for $D >0$.}
    \label{fig:training dynamics}
\end{figure}

\textbf{Relevance of symmetry breaking}. The phase transitions we studied also involve symmetry breaking. This can be seen directly from the effective landscape in Eq.~\eqref{eq: general effective loss}. The loss is unaltered as one flip the sign of $b$, and therefore the loss is symmetric in $b$. Figure~\ref{fig:training dynamics} illustrates the effect and importance of symmetry breaking on the gradient descent dynamics. Additionally, this observation may also provide an alternative venue for studying general symmetry-breaking dynamics because the computation with neural networks is both accurate and efficient. 

\textbf{Mean-Field Analysis}. The crux of our theory can be understood by applying a simplified ``mean-field" analysis of the loss function in Eq.~\eqref{eq: intro loss}. Let each weight matrix be approximated by a scalar $U = b_{D+1}$, $W_i=b_i$, ignore the stochasticity due to $\epsilon_i$, and let $x$ be one-dimensional. One then obtains a simplified mean-field loss:
\begin{equation}
    \E_x\left[\left(c_0 x\prod_{i=1}^{D+1}  b_i - y\right)^2 \right] + \gamma \sum_{i=1}^D c_i b_i^2,
\end{equation}
where $c_i$'s are constants. The first term can be interpreted as a special type of $(D+1)$-body interaction. We now perform a second mean-field approximation, where all the $b_i$ take the same value $b$:
\begin{equation}
    \ell \propto  c_0' \E[x^2]b^{2D+2}  - c_1' \E[xy] b^{D +1 }  + \gamma c_2' b^2 + const.
\end{equation}
Here, $c_0'$, $c_1'$ and $c_2'$ are structural constants, only depending on the model (depth, width, etc). The first and the third terms monotonically increase in $b$, encouraging a smaller $b$. The second term monotonically decreases in $b^{D+1} \E[xy]$, encouraging a positive correlation between $b$ and the feature $\E[xy]$. The leading and lowest-order terms regularize the model, while the intermediate term characterizes learning. For $D=0$, the loss is quadratic and has no transition. For $D=1$, the loss is identical to the Landau free energy, and a phase transition occurs when the second-order term flips sign: $c_2'\gamma = c_1'\E[xy]$. For $D>1$, the origin is always a local minimum, dominated by the quadratic term. This leads to a first-order phase transition. When $D\to \infty$, the leading terms become discontinuous in $b$, and one obtains a zeroth-order phase transition. This simple analysis highlights one important distinction between physics and machine learning: in physics, the most common type of interaction is a two-body interaction, whereas, for machine learning, the common interaction is many-body and tends to infinite-body as $D$ increases.

One implication is that $L_2$ regularization may be too strong for deep learning because it creates a trivial phase. Our result also suggests a way to avoid the trivial phase. Instead of regularizing by $\gamma ||w||_2^2$, one might consider $\gamma ||w||_2^{d+2}$, which is the lowest-order regularization that does not lead to a trivial phase. The effectiveness of this suggested method is confirmed in Supplementary Section~\ref{app: removing trivial phase}.

\textbf{Posterior Collapse in Bayesian Deep Learning}. Our results also identify an origin of the well-known problem posterior collapse problem in Bayesian deep learning. Posterior collapse refers to the learning failure where the learned posterior distribution coincides with the prior, and so no learning has happened even after training \cite{dai2019diagnosing, alemi2018fixing, lucas2019dont}. Our results offer a direct explanation for this posterior collapse problem. In the Bayesian interpretation, the training loss in Eq.~\eqref{eq: intro loss} is the exact negative log posterior, and the trivial phase exactly corresponds to the posterior collapse: the global minimum of the loss is identical to the global maximum of the prior term. Our results thus imply that (1) posterior collapse is a unique problem of deep learning because it does not occur in shallow models, and (2) posterior collapse happens as a direct consequence of the competition between the prior and the likelihood. This means that it is not a good idea to assume a Gaussian prior for the deep neural network models. The suggested fix also leads to a clean and Bayesian-principled solution to the posterior collapse problem by using a prior $\log p(w)\propto-||w||_2^{D+2}$. 

\section*{Discussion}

The striking similarity between phase transitions in neural networks and statistical-physics phase transitions lends a great impetus to a more thorough investigation of deep learning through the lens of thermodynamics and statistical physics. We now outline a few major future steps:
\begin{enumerate}
    \item Instead of classification by analyticity, can we classify neural networks by symmetry and topological invariants?
    \item What are other possible phases for a nonlinear network? Does a new phase emerge? 
    \item Can we find any analogy of other thermodynamic quantities such as volume and pressure? More broadly, can we establish thermodynamics for deep learning? 
    \item Can we utilize the latent heat picture to devise better algorithms for escaping local minima in deep learning? 
\end{enumerate}
This work shows that the Ehrenfest classification of phase transitions aligns precisely with the number of layers in deep neural networks. We believe that the statistical-physics approach to deep learning will bring about fruitful developments in both fields of statistical physics and deep learning.

\bibliographystyle{unsrt}

\clearpage
\appendix

\section{Additional Experiments}
\subsection{Sensitivity to the Initial Condition}\label{app: sensitivity to the initial condition}
\begin{figure}[t!]
     \centering
     \includegraphics[width=0.35\linewidth]{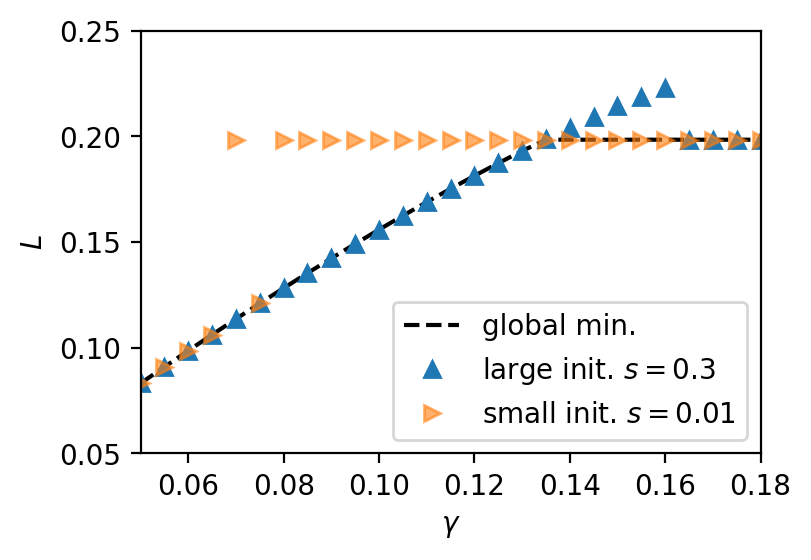}
        \vspace{-1em}
     \caption{Sensitivity of the obtained solution to the initialization of the model. We initialize the model around zero with standard deviation $s$. The experiment shows that a larger initialization variance ($s=0.3$) affords a preference of the nontrivial solution over the trivial one, while a smaller initialization leads to the opposite preference.}
     \label{fig:initial condition}
 \end{figure}

Our result suggests that the learning of a deeper network is quite sensitive to the initialization schemes we use. In particular, for $D>1$, some initialization schemes converge to the trivial solutions more easily, while others converge to the nontrivial solution more easily. Figure~\ref{fig:initial condition} plots the converged loss of a $D=2$ model for two types of initialization: (a) larger initialization, where the parameters are initialized around zero with the standard deviation $s=0.3$ and (b) small initialization with $s=0.01$. The value of $s$ is thus equal to the expected norm of the model at initialization, and a small $s$ means that it is initialized closer to the trivial phase and a larger $s$ means that it is initialized closer to the learning phase. We see that across a wide range of $\gamma$, one of the initialization schemes gets stuck in a local minimum and does not converge to the global minimum. In light of the latent heat picture, the reason for the sensitivity to initial states is clear: one needs to inject additional energy to the system to leave the meta-stable state; otherwise, the system may become stuck for a very long time. The existing initialization methods are predominantly data-dependent. However, our result (also see \cite{ziyin2022exact}) suggests that the size of the trivial minimum is data-dependent, and our result thus highlights the importance of designing data-dependent initialization methods in deep learning.

\subsection{Removing the Trivial Phase}\label{app: removing trivial phase}
\begin{figure}[t!]
     \centering

    \includegraphics[width=0.35\linewidth]{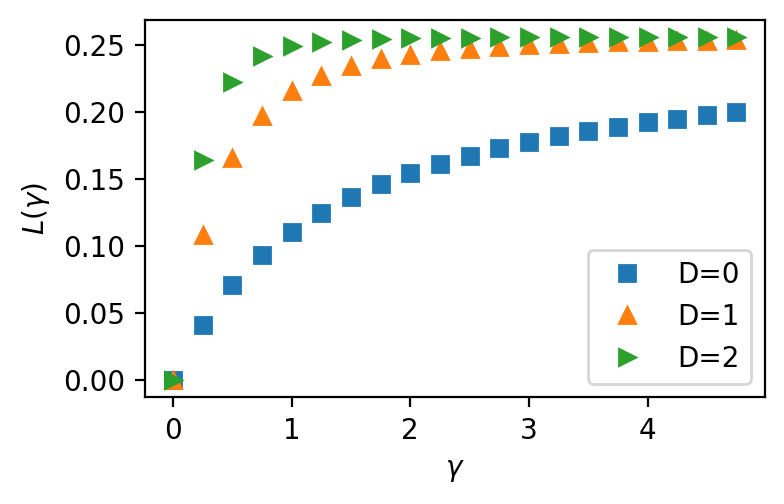}
    \includegraphics[width=0.35\linewidth]{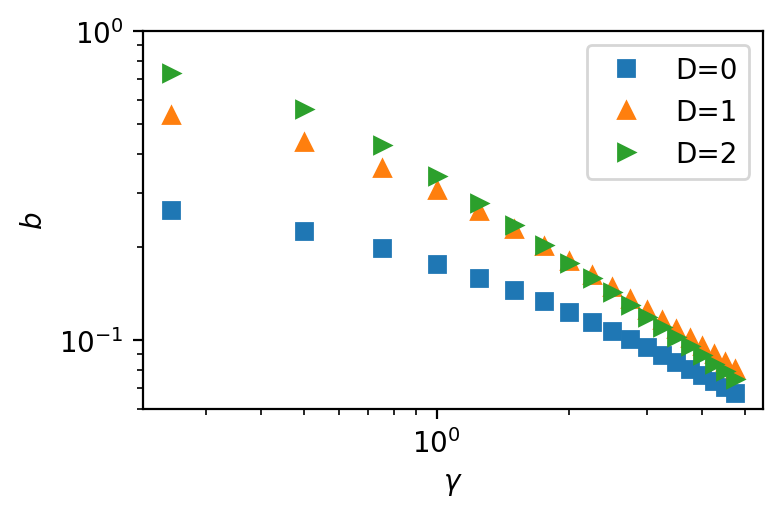}
    \vspace{-1em}
    \caption{The training loss $L(\gamma)$ (left) and the model norm $b$ (right) when we train with a regularization term of the form $\gamma||w||^{D+2}$, which is a theoretically justified fix to the trivial learning problem. We see that the trivial phase disappears under this regularization.}
    \label{fig:fixing trivial learning}
 \end{figure}

We also explore our suggested fix to the trivial learning problem. Here, instead of regularization the model by $\gamma ||w||_2^2$, we regularize the model by $\gamma ||w||_2^{D+2}$. The training loss and the model norm $b$ are plotted in Figure~\ref{fig:fixing trivial learning}. We find that the trivial phase now completely disappears even if we go to very high $\gamma$. However, we note that this fix only removes the local maximum at zero, but zero remains a saddle point from which it takes the system a long time to escape.

\begin{figure}[t!]
    \centering
    \includegraphics[width=0.32\linewidth]{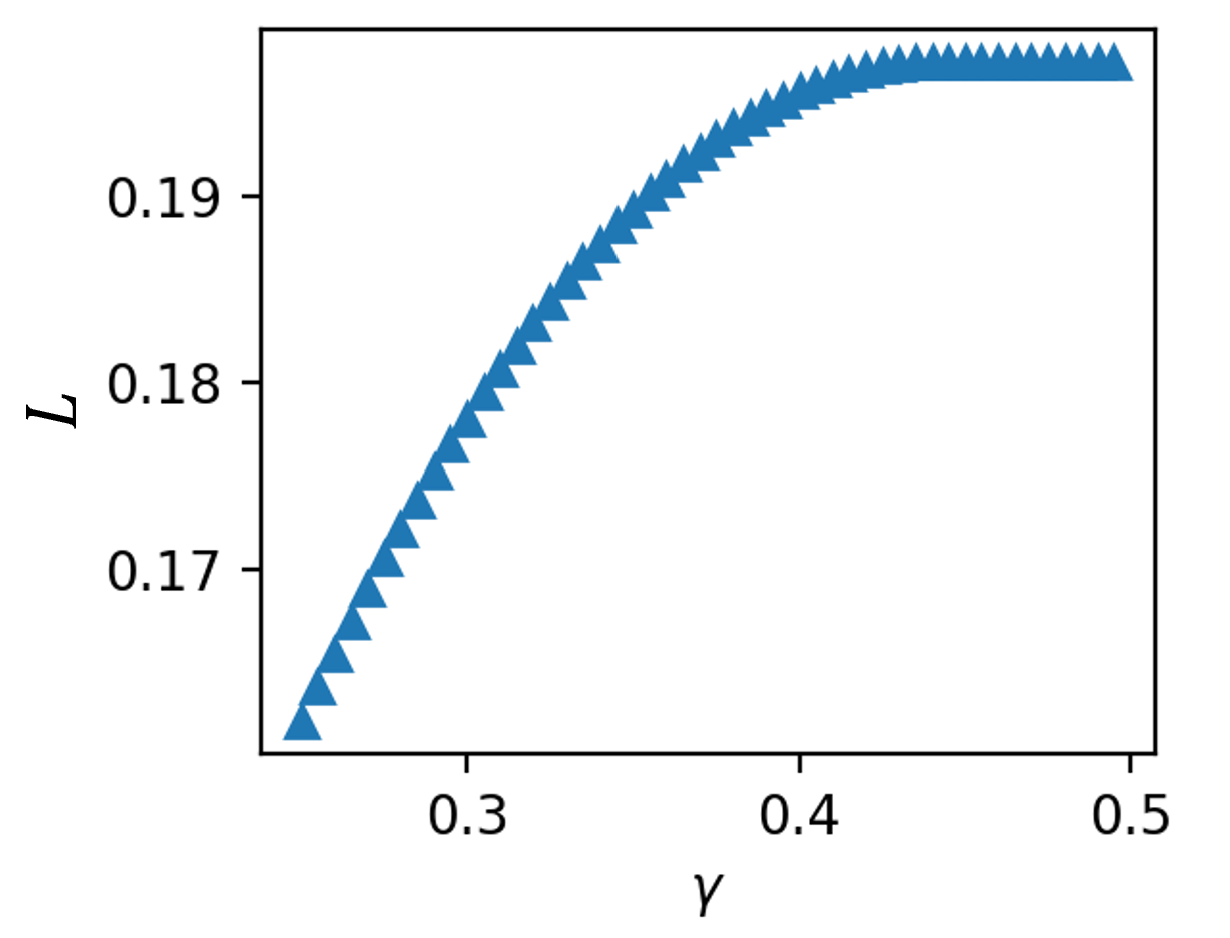}
    \includegraphics[width=0.32\linewidth]{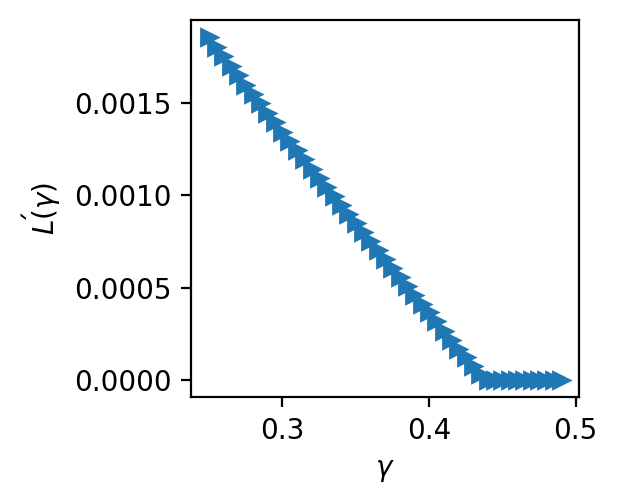}
    \includegraphics[width=0.28\linewidth]{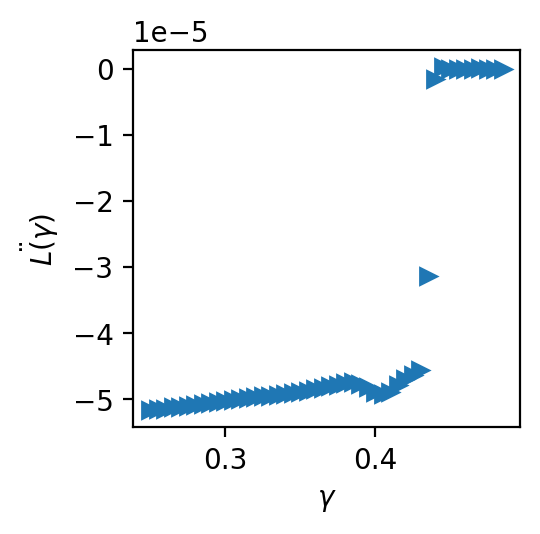}
    
    \includegraphics[width=0.32\linewidth]{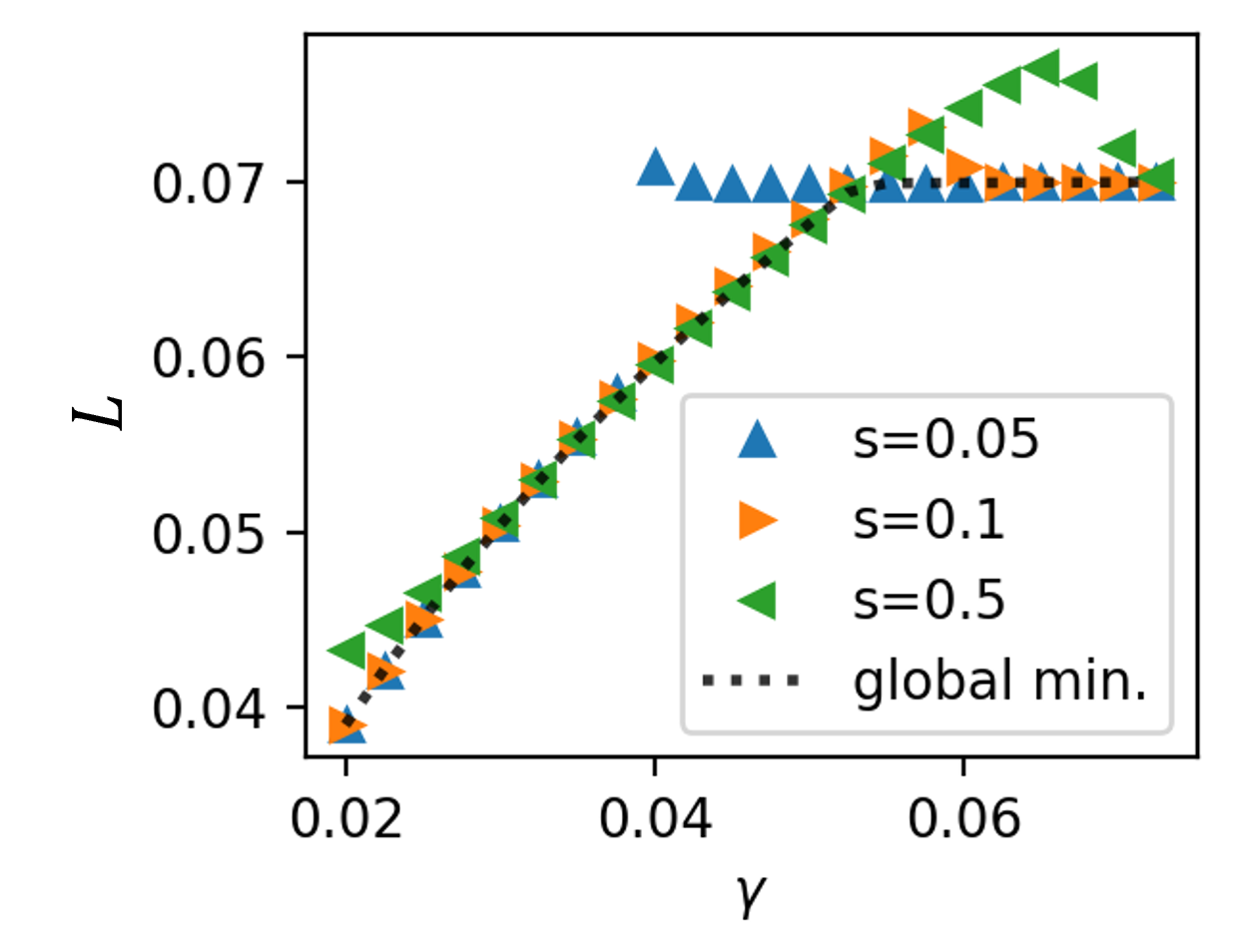}
    \includegraphics[width=0.32\linewidth]{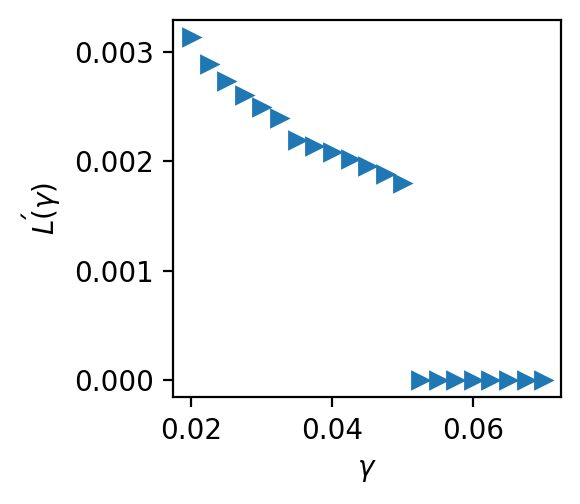}

    \caption{Phase transition of a fully connected tanh network. \textbf{Top} row shows the case of $D=1$, exhibiting a second-order phase transition: the training loss $L(\gamma)$ (left), first derivative (middle), and the second derivative (right). \textbf{Bottom} row shows the case of $D=2$, exhibiting a first-order phase transition: the training loss $L(\gamma)$ (left) and first derivative $L'(\gamma)$ (middle). For $D=2$, we initialize the model with three initialization at different scales and use the minimum of the respective loss values as an empirical estimate of the actual global minimum.}

    \label{fig:nonlinear model}
\end{figure}

\subsection{Nonlinear Networks}\label{app: nonlinear nets}
We expect our theory to also apply to deep nonlinear networks that can be locally approximated by linear net at the origin, e.g., a network with tanh activations. As shown in Figure~\ref{fig:nonlinear model}, the data shows that a tanh net also features a second-order phase transition for $D=1$ and a first-order phase transition for $D=2$.

One notable exception that our theory may not apply is the networks with the ReLU activation because these networks are not differentiable at zero (i.e., in the trivial phase). However, there are smoother (and empirically better) alternatives to ReLU, such as the swish activation function, to which the present theory should also be relevant.

\clearpage

\section{Main Results}\label{app: theory}

\subsection{Theorem Statements}
For a simple ridge linear regression, the minimization objective is 
\begin{equation}\label{eq: linear regression main loss}
    \ell(W) = \E_x  \left(\sum_{i} W_{i} x_i  - y\right)^2  + \gamma ||W||^2.
\end{equation}
\begin{theorem}\label{theo: depth 0}
    There is no phase transition in any hyperparameter {\rm($\gamma,\ A_0,\ E[xy],\ E[y^2]$)} in a simple ridge linear regression for any $\gamma \in (0,\infty)$.
\end{theorem}

The following result shows that for a finite depth, $L(\gamma)$ must be continuous in $\gamma$. 
\begin{theorem}\label{theo: no zeroth order}
    For any finite $D > 0$ and $\gamma \in [0,\infty)$, $L(\gamma)$ has no zeroth-order phase transition with respect to $\gamma$.
\end{theorem}
Note that this theorem allows the weight decay parameter to be $0$, and so our results also extend to the case when there is no weight decay.

The following theorem shows that there exists order parameters describing any phase transition induced by the weight decay parameter in Eq.~\eqref{eq: intro loss}.
\begin{theorem}\label{theo: existence of order parameter}
     Let $b= ||U||/d_u$, and let
    \begin{equation}
        \bar{\ell}(b, \gamma) := -\sum_i \frac{d_0^{2D}b^{2D} \E[x'y]_i^2 }{d_0^D (\sigma^2+d_0)^D a_i b^{2D} + \gamma}  + \E_x[y^2] +  \gamma D d_0^2 b^2.
    \end{equation} 
    Then, $b$ is an order parameter of Eq.~\eqref{eq: intro loss} for the effective loss $\bar{\ell}$.
\end{theorem}

Here, the norm of the last layer is referred to as the order parameter. The meaning of this choice should be clear. The norm of the last layer is zero if and only if all weights of the last layer is zero, and the model is a trivial model. The model can only learn something when the order parameter is nonzero. Additionally, we note that the choice of the order parameter is not unique and there are other choices for the order parameter (e.g., the norm of any other layer, or the sum of the norms of all layers). 

The following theorem shows that when $D=1$ in Eq.~\eqref{eq: intro loss}, there is a second-order phase transition with respect to $\gamma$.

\begin{theorem}\label{theo: depth 1}
    Equation~\eqref{eq: intro loss} has the second-order phase transition between the trivial and feature learning phases at\footnote{When the two layers have different regularization strengths $\gamma_u$ and $\gamma_w$, one can show that the phase transition occurs precisely at $\sqrt{\gamma_u \gamma_w} = ||\E[xy]||$.}
    \begin{equation}
        \gamma = ||\E[xy]||.
    \end{equation}
\end{theorem}

Now, we show that for $D\geq 2$, there is a first-order phase transition.
\begin{theorem}\label{theo: depth 1+}
    Let $D \geq 2$. There exists a $\gamma^*>0$ such that the loss function Eq.~\eqref{eq: intro loss} has the first-order phase transition between the trivial and feature learning phases at $\gamma= \gamma^*$.
\end{theorem}

\begin{theorem}\label{theo: infinite D}
    Let $L^{(D)}(\gamma)$ denote the loss function for a fixed depth $D$ as a function of $\gamma$. Then, for $\gamma\in[0, \infty)$ and some constant $r$,
    \begin{equation}
        L^{(D)} (\gamma) \to \begin{cases}
            r &\text{if $\gamma=0$};\\
            \E[y^2] &\text{otherwise}.
        \end{cases}
    \end{equation}
\end{theorem}

The constant $r$ is, in general, not equal to $\E[y^2]$. For example, in the limit $\sigma\to 0$, $r$ converges to the loss value of a simple linear regression, which is not equal to $\E[y^2]$ as long as $\E[xy] \neq 0$.

\subsection{Proof of Theorem~\ref{theo: depth 0}}
\textit{Proof}. The global minimum of Eq.~\eqref{eq: linear regression main loss} is
\begin{equation}
    W_* =  (A_0 + \gamma I )^{-1}E[xy].
\end{equation}
The loss of the global minimum is thus
\begin{align}
    L &=\E_x  \left(\sum_{i} W_{i} x_i  - y\right)^2  + \gamma ||W||^2\\
    &= W^T A_0 W - 2W^T \E[xy] + \E[y^2] + \gamma ||W||^2\\
    &=\E[xy]^T \frac{A_0}{(A_0 + \gamma I)^2}\E[xy] - 2\E[xy]^T \frac{1}{A_0 + \gamma I }\E[xy] + \E[y^2] + \gamma \E[xy]^T \frac{1}{(A_0 + \gamma I)^2}\E[xy]\\
    &=- \E[xy]^T(A_0 + \gamma I)^{-1} \E[xy]  + \E[y^2],
\end{align}
which is infinitely differentiable for any $\gamma \in (0,\infty)$ (note that $A_0$ is always positive semi-definite by definition). $\square$
\subsection{Proof of Theorem~\ref{theo: no zeroth order}}
\textit{Proof}. For any fixed and bounded $w$, $\ell(w,\gamma)$ is continuous in $\gamma$. Moreover, $\ell(w,\gamma)$ is a monotonically increasing function of $\gamma$. This implies that $L(\gamma)$ is also an increasing function of $\gamma$ (but may not be strictly increasing). 

We now prove by contradiction. We first show that $L(\gamma)$ is left-continuous. Suppose that for some $D$, $L(\gamma)$ is not left-continuous in $\gamma$ at some $\gamma^*$. By definition, we have
\begin{equation}
    L(\gamma^*-\epsilon) = \min_{w} \ell(w, \gamma^*-\epsilon) := \ell(w', \gamma^*-\epsilon),
\end{equation}
where $w'$ is one of the (potentially many) global minima of $L(\gamma^* -\epsilon)$. Since $L(\gamma)$ is not left-continuous by assumption, there exists $\delta >0$ such that for any $\epsilon > 0$, 
\begin{equation}
     L(\gamma^* - \epsilon) < L (\gamma^*) -\delta,
\end{equation}
which implies that
\begin{equation}\label{eq: proof ineq 1}
    \ell(w', \gamma^* - \epsilon )  =  L(\gamma^* - \epsilon)< L (\gamma^*) -\delta \leq \ell(w', \gamma^*) - \delta.
\end{equation}
Namely, the left-discontinuity implies that for all $\epsilon >0$,
\begin{equation}
    \ell(w', \gamma^* - \epsilon ) \leq \ell(w', \gamma^*) - \delta.
\end{equation}
However, by definition of $\ell(w, \gamma)$, we have
\begin{equation}
    \ell(w, \gamma) - \ell(w, \gamma-\epsilon) = \epsilon ||w||^2.
\end{equation}
Thus, by choosing $\epsilon < \delta/||w||^2$, the relation in \eqref{eq: proof ineq 1} is violated. Thus, $L(\gamma)$ must be left-continuous. 

In a similar manner, we can prove that $L$ is right-continuous. Suppose that for some $D$, $L(\gamma)$ is not right-continuous in $\gamma$ at some $\gamma^*$. Let $\gamma>0$. By definition, we have
\begin{equation}
    L(\gamma^*+\epsilon) = \min_{w} \ell(w, \gamma^*+\epsilon) := \ell(w', \gamma^*+\epsilon),
\end{equation}
where $w'$ is one of the (potentially many) global minima of $L(\gamma^* +\epsilon)$. Since $L(\gamma)$ is not right-continuous by assumption, there exists $\delta >0$ such that for any $\epsilon > 0$, 
\begin{equation}
     L(\gamma^* + \epsilon) > L (\gamma^*) + \delta,
\end{equation}
which implies that
\begin{equation}\label{eq: proof ineq 2}
    \ell(w', \gamma^* + \epsilon )  =  L(\gamma^* + \epsilon) > L (\gamma^*) +\delta \geq \ell(w', \gamma^*) + \delta.
\end{equation}
Namely, the right-discontinuity implies that for all $\epsilon >0$,
\begin{equation}
    \ell(w', \gamma^* + \epsilon ) \geq \ell(w', \gamma^*) + \delta.
\end{equation}
However, by definition of $\ell(w, \gamma)$, we have
\begin{equation}
    \ell(w, \gamma+\epsilon) - \ell(w, \gamma) = \epsilon ||w||^2.
\end{equation}
Thus, by choosing $\epsilon < \delta/||w||^2$, the relation in \eqref{eq: proof ineq 2} is violated. Thus, $L(\gamma)$ must be right-continuous. 

Therefore, $L(\gamma)$ is continuous for all $\gamma > 0$. By definition, this means that there is no zeroth-order phase transition in $\gamma$ for $L$. Additionally, note that the above proof does not require $\gamma \neq 0$, and so we have also shown that $L(\gamma)$ is right-continuous at $\gamma=0$. 
$\square$

\subsection{Proof of Theorem~\ref{theo: existence of order parameter}}
\textit{Proof}. By Theorem 3 of Ref.~\cite{ziyin2022exact}, any global minimum of Eq.~\eqref{eq: intro loss} is given by the following set of equations for some $b\geq 0$:
\begin{equation}
\label{eq: homogeneous special solution D layer b}
    \begin{cases}
        U =  \sqrt{d_0} b \mathbf{r}_{D};\\
        W^{(i)} = b \mathbf{r}_i\mathbf{r}_{i-1}^T;\\
        W^{(1)} = 
        \mathbf{r}_1\E[xy]^T d_0^{D - \frac{1}{2}} b^D \left[d_0^D ( \sigma^2 + d_0)^D b^{2D} A_0 + \gamma \right]^{-1},
    \end{cases}
\end{equation}
where $\mathbf{r}_i=(\pm 1, ...,\pm 1)$ is an arbitrary vertex of a $d_i$-dimensional hypercube for all $i$. Therefore, the global minimum must lie on a one-dimensional space indexed by $b\in [0, \infty)$. Let $f(x)$ specify the model as
\begin{equation}
    f(x):=\sum_{i,i_1,i_2,...,i_D}^{d,d_1,d_2,...d_D} U_{i_D} \epsilon^{(D)}_{i_{D}}...\epsilon_{i_2}^{(2)} W_{i_2i_1}^{(2)}\epsilon_{i_1}^{(1)} W_{i_1i}^{(1)} x,
\end{equation}
and let $\eta$ denote the set of all random noises $\epsilon_i$.

Substituting Eq.~\eqref{eq: homogeneous special solution D layer b} in Eq.~\eqref{eq: intro loss}, one finds that within this subspace, the loss function can be written as
\begin{align}
    \ell(w, \gamma) &= \E_x\E_{\eta}(f(x) -y)^2 + L_2\ {\rm reg.}\\
    &=  \E_{x,\eta}[f(x)^2] - 2\E_{x,\eta}[yf(x)] + \E_x[y^2] + L_2\ {\rm reg.} \\
    & =  \sum_i \frac{d_0^{3D}(\sigma^2 + d_0)^D b^{4D} a_i \E[x'y]_i^2 }{[d_0^D (\sigma^2+d_0)^D a_i b^{2D} + \gamma]^2} - 2  \sum_i\frac{d_0^{2D} b^{2D} \E[x'y]^2_i}{d_0^D (\sigma^2+d_0)^D a_i b^{2D} + \gamma} + \E_x[y^2] +  L_2\ {\rm reg.},
\end{align}
where the $L_2\ {\rm reg.}$ term is
\begin{equation}
    L_2\ {\rm reg.} =  \gamma D d_0^2 b^2 + \gamma \sum_i \frac{d_0^{2D} b^{2D}  ||\E[x'y]_i||^2}{[d_0^D (\sigma^2 + d_0)^D b^{2D} a_i + \gamma]^2}.
\end{equation}
Combining terms, we can simplify the expression for the loss function to be
\begin{equation}
    -\sum_i \frac{d_0^{2D}b^{2D} \E[x'y]_i^2 }{[d_0^D (\sigma^2+d_0)^D a_i b^{2D} + \gamma]}  + \E_x[y^2] +  \gamma D d_0^2 b^2.
\end{equation}
We can now define the effective loss by
\begin{equation}
    \bar{\ell}(b,\gamma) := -\sum_i \frac{d_0^{2D}b^{2D} \E[x'y]_i^2 }{[d_0^D (\sigma^2+d_0)^D a_i b^{2D} + \gamma]}  + \E_x[y^2] +  \gamma D d_0^2 b^2.
\end{equation}
Then, the above argument shows that, for all $\gamma$,
\begin{equation}
    \min_{b} \bar{\ell}(b,\gamma) = \min_{w} \ell(w,\gamma).
\end{equation}
By definition~\ref{def: order parameter}, $b$ is an order parameter of $\ell$ with respect to the effective loss $\bar{\ell}(b,\gamma)$. This completes the proof. $\square$

\subsection{Two Useful Lemmas}
Before continuing the proofs, we first prove two lemmas that will simplify the following proofs significantly.
\begin{lemma}\label{lemma: first derivative}
If $L(\gamma)$ is differentiable, then for at least one of the global minima $b_*$,
\begin{equation}
    \frac{d}{d\gamma} L(\gamma) = \sum_i \frac{d_0^{2D}b_*^{2D} \E[x'y]_i^2 }{[d_0^D (\sigma^2+d_0)^D a_i b_*^{2D} + \gamma]^2}   +  D d_0^2 b_*^2 \geq 0.
\end{equation}
\end{lemma}
\textit{Proof}. Because $L$ is differentiable in $\gamma$, one can find the derivative for at least one of the global minima $b^*$
\begin{align}
    \frac{d}{d\gamma} L(\gamma) &= \frac{d}{d\gamma}  \bar{\ell}(b^*(\gamma), \gamma)\\
    &= \frac{\partial}{\partial b^*}  \bar{\ell}(b^*, \gamma) \frac{\partial b^*}{\partial \gamma}  + \frac{\partial}{\partial \gamma}  \bar{\ell}(b^*, \gamma)\\
    &= \frac{\partial}{\partial \gamma}  \bar{\ell}(b^*, \gamma)\\
    & = \sum_i \frac{d_0^{2D}b_*^{2D} \E[x'y]_i^2 }{[d_0^D (\sigma^2+d_0)^D a_i b_*^{2D} + \gamma]^2}   +  D d_0^2 b_*^2 \geq 0,
\end{align}
where we have used the optimality condition $\frac{\partial}{\partial b^*}  \bar{\ell}(b^*(\gamma), \gamma) =0$ in the second equality. $\square$

\subsection{Proof of Theorem~\ref{theo: depth 1}}
\textit{Proof}. By definition~\ref{def: phase transition}, it suffices to only prove the existence of phase transitions on the effective loss. For $D=1$, the effective loss is
\begin{equation}\label{eq: effective loss 2 layer}
    \bar{\ell}(b,\gamma) = -d_1 b^2 E[xy]^T [b^2(\sigma^2 + d_1)A + \gamma I]^{-1}E[xy] + E[y^2] + \gamma d_1 b^2.
\end{equation}
By Theorem 1 of Ref.~\cite{ziyin2022exact}, the phase transition, if exists, must occur precisely at $\gamma =||\E[xy]||$. To prove that $\gamma =||\E[xy]||$ has a second-order phase transition, we must check both its first derivative and second derivative.

When $\gamma \to ||E[xy]||$ from the right, we find that the all derivatives of $L(\gamma)$ are zero because the loss is identically equal to $\E[y^2]$. We now consider the derivative of $L$ when $\gamma \to ||E[xy]||$ from the left.

We first need to find the minimizer of Eq.~\eqref{eq: effective loss 2 layer}. Because Eq.\eqref{eq: effective loss 2 layer} is differentiable, its derivative in $b$ must be equal to $0$ at the global minimum
\begin{equation}\label{eq: proof solution condition}
    -2\gamma d_1 b \E[xy]^T [b^2(\sigma^2 + d_1)^2 A + \gamma I ]^{-2} \E[xy] + 2\gamma d_1 b = 0.
\end{equation}
Finding the minimizer $b$ is thus equivalent to finding the real roots of a high-order polynomial in $b$.  When $\gamma \geq ||\E[xy]||$, the solution is unique \cite{ziyin2022exact}:
\begin{equation}
    b^2_0 = 0,
\end{equation}
where we labeled the solution with the subscript $0$ to emphasize that this solution is also the zeroth-order term of the solution in a perturbatively small neighborhood of $\gamma = ||E[xy]||$. From this point, we define a shifted regularization strength: $\Delta := \gamma - ||\E[xy]||$. When $\Delta <0$, the condition~\eqref{eq: proof solution condition} simplifies to
\begin{equation}
    \E[xy]^T [b^2(\sigma^2 + d_1) A + \gamma I ]^{-2} \E[xy] = 1.
\end{equation}
Because the polynomial is not singular in $\Delta$, one can Taylor expand the (squared) solution $b^2$ in $\Delta$:
\begin{equation}\label{eq: proof solution expansion}
    b(\gamma)^2 = \beta_0 + \beta_1 \Delta + O (\Delta^2).
\end{equation}
We first Substitute \eqref{eq: proof solution expansion} in \eqref{eq: proof solution condition} to find\footnote{Note that alternatively, $\beta_0=0$ is implied by the no-zeroth-order transition theorem.}
\begin{equation}
    \beta_0=0.
\end{equation}
One can then again substitute Eq.~\eqref{eq: proof solution expansion} in Eq.~\eqref{eq: proof solution condition} to find $\beta_1$. To the first order in $b^2$, Eq.~\eqref{eq: proof solution condition} reads
\begin{align}
    &\frac{1}{\gamma^2} ||\E[xy]||^2 - 2b^2\frac{(\sigma^2 +d_1)}{\gamma^3} ||\E[xy]||_{A_0}^2 = 1\\
    &\Longleftrightarrow 
    - 2\beta_1 \Delta \frac{(\sigma^2 +d_1)}{\gamma^3} ||\E[xy]||_{A_0}^2 = 2\frac{\Delta}{||\E[xy]||}\\
    &\Longleftrightarrow 
    \beta_1  = - \frac{1}{(\sigma^2 + d_1)}\frac{||\E[xy]||^2}{||\E[xy]||^2_{A_0}}
\end{align}
Substituting this first-order solution to Lemma~\ref{lemma: first derivative}, we obtain that
\begin{equation}
     \frac{d}{d\gamma}L(\gamma)|_{\gamma = ||E[xy]||_{-}} \sim b_*^2 =0 = \frac{d}{d\gamma}L(\gamma)|_{\gamma = ||E[xy]||_{+}}.
\end{equation}
Thus, the first-order derivative of $L(\gamma)$ is continuous at the phase transition point. 

We now find the second-order derivative of $L(\gamma)$. To achieve this, we also need to find the second-order term of $b^2$ in $\gamma$. We expand $b^2$ as
\begin{equation}
    b(\gamma)^2 = 0 + \beta_1 \Delta + \beta_2 \Delta^2 + O (\Delta^3).
\end{equation}
To the second order in $b^2$, \eqref{eq: proof solution condition} reads
\begin{align}
    &\frac{1}{\gamma^2} ||\E[xy]||^2 - 2b^2\frac{(\sigma^2 +d_1)}{\gamma^3} ||\E[xy]||_{A_0}^2 + 3 b^4\frac{(\sigma^2+ d_1)^2}{\gamma^4} ||\E[xy]||_{A_0^2}^2 = 1\\
    &\Longleftrightarrow 
    {\gamma^2} ||\E[xy]||^2 - 2b^2{(\sigma^2 +d_1)}{\gamma} ||\E[xy]||_{A_0}^2 + 3 b^4{(\sigma^2+ d_1)^2} ||\E[xy]||_{A_0^2}^2 = \gamma^4\\
    &\Longleftrightarrow 
    {\Delta^2} E_0^2 - 2\beta_2\Delta^2{(\sigma^2 +d_1)}{E_0} E_1^2 - 2\beta_1\Delta^2{(\sigma^2 +d_1)} E_1^2 + 3 \beta_1^2 \Delta^2{(\sigma^2+ d_1)^2} E_2^2 =  6E_0^2\Delta^2 \\
    &\Longleftrightarrow \beta_2 = \frac{3\beta_1^2 (\sigma^2 +d_1)^2 E_2^2 -5E_0^2 - 2\beta_1 (\sigma^2 +d_1) E_1^2}{2(\sigma^2 + d_1) E_0 E_1^2},
\end{align}
where, from the third line, we have used the shorthand $E_0 := ||\E[xy]||$, $E_1 := ||\E[xy]||_{A_0}$, and $E_2 := ||\E[xy]||_{A_0^2}$.
Substitute in $\beta_1$, one finds that
\begin{equation}
    \beta_2 = \frac{3E_0 (E_2^2 -E_1^2)}{2(\sigma^2 + d_1) E_1^4}.
\end{equation}


This allows us to find the second derivative of $L(\gamma)$. Substituting $\beta_1$ and $\beta_2$ into Eq.~\eqref{eq: effective loss 2 layer} and expanding to the second order in $\Delta$, we obtain that
\begin{align}
    L(\gamma) &= -d_1 b^2 E[xy]^T [b^2(\sigma^2 + d_1)A + \gamma I]^{-1}E[xy] + E[y^2] + \gamma d_1 b^2\\
    &= -d_1 (\beta_1 \Delta + \beta_2 \Delta^2) \E[xy]^T[(\beta_1 \Delta + \beta_2 \Delta^2) (\sigma^2 + d_1) A_0 + \gamma I]^{-1} \E[xy] + \gamma d_1 (\beta_1 \Delta + \beta_2 \Delta).
\end{align}
At the critical point,
\begin{align}
    \frac{d^2}{d\gamma^2} L(\gamma)|_{\gamma = ||\E[xy]||_{-}} &= -d_1 \beta_2  E_0 + d_1  \beta_1^2 (\sigma^2 + d_1) \frac{E_1^2}{E_0^2} + d_1 \beta_1  + d_1 \beta_1 + d_1 \beta_2 E_0  \\
    &= 2d_1 \beta_1 +  d_1  \beta_1^2 (\sigma^2 + d_1) \frac{E_1^2}{E_0^2}\\
    &= d_1 \beta_1\\
    &= -\frac{d_1}{\sigma^2 + d_1} \frac{||\E[xy]||^2}{||\E[xy]||^2_{A_0}}.
\end{align}
Notably, the second derivative of $L$ from the left is only dependent on $\beta_1$ and not on $\beta_2$.


\begin{equation}
    \frac{d^2}{d\gamma^2} L(\gamma)|_{\gamma = ||\E[xy]||_{-}} = -\frac{d_1}{\sigma^2 + d_1} \frac{||\E[xy]||^2}{||\E[xy]||^2_{A_0}} <0.
\end{equation}
Thus, the second derivative of $L(\gamma)$ is discontinuous at $\gamma=||\E[xy]||$. This completes the proof. $\square$

\begin{remark}
    Note that the proof suggests that close to the critical point, $b\sim \sqrt{\Delta}$, in agreement with the Landau theory.
\end{remark}

\subsection{Proof of Theorem~\ref{theo: depth 1+}}

\textit{Proof}. By definition, it suffices to show that $\frac{d}{d\gamma} L(\gamma)$ is not continuous. We prove by contradiction. Suppose that $\frac{d}{d\gamma} L(\gamma)$ is everywhere continuous on $\gamma \in (0,\infty)$. Then, by Lemma~\ref{lemma: first derivative}, one can find the derivative for at least one of the global minima $b^*$
\begin{align}
    \frac{d}{d\gamma} L(\gamma) = \sum_i \frac{d_0^{2D}b_*^{2D} \E[x'y]_i^2 }{[d_0^D (\sigma^2+d_0)^D a_i b_*^{2D} + \gamma]^2}   +  \gamma D d_0^2 b_*^2 \geq 0.
\end{align}
Both terms in the last line are nonnegative, and so one necessary condition for $\frac{d}{d\gamma} L(\gamma)$ to be continuous is that both of these two terms are continuous in $\gamma$. 

In particular, one necessary condition is that $\gamma D d_0^2 b_*^2$ is continuous in $\gamma$. By Proposition 3 of Ref.~\cite{ziyin2022exact}, there exist constants $c_0$, $c_1$ such that $0 < c_0 \leq c_1$, and
\begin{equation}
    \begin{cases}
        b_* = 0 & \text{if }\gamma < c_0;\\
        b_* >0, & \text{if }\gamma > c_1.
    \end{cases}
\end{equation}

Additionally, if $b_* >0$, $b_*$ must be lower-bounded by some nonzero value \cite{ziyin2022exact}:
\begin{equation}
    b_* \geq \frac{1}{d_0}\left( \frac{\gamma}{||\E[xy]||} \right)^{\frac{1}{D-1}} > \frac{1}{d_0}\left( \frac{c_1}{||\E[xy]||} \right)^{\frac{1}{D-1}} > 0.
\end{equation}
Therefore, for any $D>1$, $b_*(\gamma)$ must have a discontinuous jump from $0$ to a value larger than $\frac{1}{d_0}\left( \frac{c_0}{||\E[xy]||} \right)^{\frac{1}{D-1}}$, and cannot be continuous. This, in turn, implies that $\frac{d}{d\gamma} L(\gamma)$ jumps from zero to a nonzero value and cannot be continuous. This completes the proof. $\square$

\subsection{Proof of Theorem~\ref{theo: infinite D}}
\textit{Proof}. It suffices to show that a nonzero global minimum cannot exist at a sufficiently large $D$, when one fixes $\gamma$. By Proposition 3 of Ref.~\cite{ziyin2022exact}, when $\gamma>0$, any nonzero global minimum must obey the following two inequalities:
\begin{equation}\label{eq: solution upper bound}
    \frac{1}{d_0}\left[\frac{\gamma}{||\E[xy]||}\right]^{\frac{1}{D-1}} \leq b^* \leq \left[\frac{||\E[xy]||}{ d_0(\sigma^2 + d_0)^D a_{\rm max}}\right]^{\frac{1}{D+1}},
\end{equation}
where $a_{\rm max}$ is the largest eigenvalue of $A_0$. In the limit $D\to \infty$, the lower bound becomes
\begin{equation}
    \frac{1}{d_0}\left[\frac{\gamma}{||\E[xy]||}\right]^{\frac{1}{D-1}}  \to \frac{1}{d_0}.
\end{equation}
The upper bound becomes 
\begin{equation}
    \left[\frac{||\E[xy]||}{ d_0(\sigma^2 + d_0)^D a_{\rm max}}\right]^{\frac{1}{D+1}} \to \frac{1}{\sigma^2 + d_0 }.
\end{equation}
But for any $\sigma^2 > 0$, $\frac{1}{d_0} < \frac{1}{\sigma^2 + d_0 }$. Thus, the set of such $b^*$ is empty. 

On the other hand, when $\gamma=0$, the global minimizer has been found in Ref.~\cite{mianjy2019dropout} and is nonzero, which implies that $L(0)<\E[y^2]$. This means that $L(\gamma)$ is not continuous at $0$. This completes the proof. $\square$

\end{document}